\documentclass[10pt, a4paper]{article}

\usepackage{hyperref}

\usepackage{lrec}
\usepackage{graphicx}
\usepackage{tabularx}
\usepackage{soul}

\usepackage[dvipsnames]{xcolor}
\definecolor{mygray}{rgb}{0.4, 0.4, 0.4}

\usepackage{titlesec}
\titleformat{\section}{\normalfont\large\bf\center}{\thesection.}{1em}{}
\titleformat{\subsection}{\normalfont\SmallTitleFont\bf\raggedright}{\thesubsection.}{1em}{}
\titleformat{\subsubsection}{\normalfont\normalsize\bf\raggedright}{\thesubsubsection.}{1em}{}
\renewcommand\thesection{\arabic{section}}
\renewcommand\thesubsection{\thesection.\arabic{subsection}}
\renewcommand\thesubsubsection{\thesubsection.\arabic{subsubsection}}
\usepackage{float}
\usepackage{subfig}

\usepackage{placeins}

\usepackage{multirow}
\usepackage{multicol}
\usepackage{epstopdf}
\usepackage[utf8]{inputenc}

\usepackage{hyperref}
\usepackage{xstring}

\RequirePackage{color}

\usepackage{framed}

\title{Stress Test Evaluation of Transformer-based Models\\ in Natural Language Understanding Tasks}

\name{Carlos Aspillaga\textsuperscript{*}\thanks{* Equal contribution, listing order is random.}$^\dagger$, Andr\'es Carvallo$^{\dagger \ddagger}$, Vladimir Araujo\textsuperscript{*}$^{\dagger \ddagger}$}

\address{$^\dagger$Pontificia Universidad Cat\'olica de Chile, Santiago, Chile \\
         $^\ddagger$IMFD, Santiago, Chile \\
         \{cjaspill, afcarvallo, vgaraujo\}@uc.cl\\}

\abstract{There has been significant progress in recent years in the field of Natural Language Processing thanks to the introduction of the Transformer architecture. Current state-of-the-art models, via a large number of parameters and pre-training on massive text corpus, have shown impressive results on several downstream tasks. Many researchers have studied previous (non-Transformer) models to understand their actual behavior under different scenarios, showing that these models are taking advantage of clues or failures of datasets and that slight perturbations on the input data can severely reduce their performance. In contrast, recent models have not been systematically tested with adversarial-examples in order to show their robustness under severe stress conditions. For that reason, this work evaluates three Transformer-based models (RoBERTa, XLNet, and BERT) in Natural Language Inference (NLI) and Question Answering (QA) tasks to know if they are more robust or if they have the same flaws as their predecessors. As a result, our experiments reveal that RoBERTa, XLNet and BERT are more robust than recurrent neural network models to stress tests for both NLI and QA tasks. Nevertheless, they are still very fragile and demonstrate various unexpected behaviors, thus revealing that there is still room for future improvement in this field.\\
\newline 
\Keywords{adversarial evaluation, stress tests, natural language inference, natural language understanding, question answering}}

\begin{document}

\maketitleabstract

\section{Introduction}
Deep learning has allowed for solving several problems related to natural language processing (NLP), even outperforming human performance in some tasks, such as multi-label classification \cite{tsoumakas2007multi}, document screening \cite{carvallocomparing}, named entity recognition \cite{nadeau2007survey}, among others. However, previous research has shown that neural networks are powerful enough to memorize the training data, which limits their ability to generalize or to really understand the tasks they are dealing with \cite{45820}. %Moreover, some recent studies propose evaluation scenarios in various natural language understanding (NLU) tasks.

One way to test NLP models is by using adversarial tests, which implies an intentional perturbation of the input sentence to confuse a model into making wrong predictions. This methodology has shown that models are still weak \cite{belinkov2018synthetic,iyyer-etal-2018-adversarial,ribeiro-etal-2018-semantically,ebrahimi-etal-2018-adversarial}. Other researchers have also shown that language models can ``falsely" solve the task. In other words, they might be taking advantage of dataset failures or artifacts on the input sentences in order to guess the answer \cite{gururangan-etal-2018-annotation,agrawal-etal-2016-analyzing,levy-etal-2015-supervised}. These evaluations, also known as ``stress tests", have been performed on classic models based on recurrent networks (RNN). However, Transformer-based models such as RoBERTa \cite{liu2019roberta}, XLNet \cite{yang2019xlnet} and BERT \cite{devlin2018bert}, which are state-of-the-art for NLU tasks, have not been systematically evaluated under severe stress conditions. Only BERT has been tested with similar objectives as ours \cite{hsieh-etal-2019-robustness,jin2019bert,niven2019probing}, but not in a systematic way as here nor in the same scenarios. 

In this work, we focus on three language models based on the state-of-the-art Transformer architecture (RoBERTa, XLNet and BERT), with the aim of carrying out a stress test evaluation on two natural language understanding (NLU) tasks. On the one hand, Natural Language Inference (NLI), also known as recognizing textual entailment (RTE) which consists of finding semantic relations between a premise sentence and an associated hypothesis, by \textbf{classifying} if they are entailed, in contradiction or in neutral relationship. On the other hand, we apply stress tests on a question-answering (QA) task, also known as machine reading comprehension (MRC) which consists of \textbf{predicting} the answer to a question given a paragraph.

The evaluation of the NLI task was performed using the MultiNLI dataset  \cite{N18-1101} following the methodology of \newcite{naik18coling}. For the QA task we used the SQuAD dataset \cite{rajpurkar-etal-2016-squad} and adversarial techniques introduced by \newcite{jia-liang-2017-adversarial}. We also developed a new adversarial dataset for SQuAD, using techniques inspired on \newcite{belinkov2018synthetic}\footnote{we released the dataset at \url{https://github.com/caspillaga/noisy-squad}}. 

All our test procedures try to prove the strength of the models, by distracting, confusing or proving their competence. Experiments show that all models are affected by stress tests, but on Transformer-based models, the adversaries have smaller impact compared to previous models based on RNNs. This behavior could be explained by the large number of parameters and their prior training. Nevertheless, in this work we not only measure the impact on performance of various adversarial or noisy conditions, but also reveal that in some cases the state-of-the-art models behave in strange and unexpected ways.

We provide detailed quantitative analysis on all the performed tests, and in some cases we report representative examples via inspection of the attention matrices that these models produce during inference when tested under adversarial test scenarios.

%The remainder of this paper is organized as follows: Section~\ref{NLU} presents language models based on the Transformer architecture; Section~\ref{NLI} and Section~\ref{QA} define the tasks to be evaluated with the stress test; Section~\ref{experiments} shows the stress tests performed for each of the tasks; Section~\ref{RW} introduces the review of the related work and discussion about them; and Section~\ref{conclusions} synthesizes the main conclusions.

\section{Transformer for Natural Language Understanding}
\label{NLU}
The Transformer \cite{vaswani2017attention} is a deep learning architecture originally proposed for neural machine translation applications. The main idea behind this model is the multi-head self-attention, the ability to attend to different parts and aspects of the input sequence to compute a contextual representation of it, at increasing levels of abstraction (layers). This architecture allows surpassing long-term dependency problems that are common on Recurrent Neural Networks (RNN) models, and adding the possibility of being highly parallelizable.

Early works such as GPT \cite{radford2018} and BERT \cite{devlin2018bert} proposed variants of the Transformer architecture for language modeling \cite{NIPS2000_1839}. These works show that the representations learned on large-scale language modeling datasets are effective for downstream sentence-level tasks (i.e. NLI) and token-level tasks (i.e. QA) via fine-tuning. However, compared to RNNs, no systematic evaluation of robustness and failure modes for these kind of models (specially the most recent variants) have been performed in previous works.

In this work, we evaluate three state-of-the-art models on their large version: BERT \cite{devlin2018bert}, which was the first model to introduce bidirectional representation in the Transformer encoder and masked modeling, XLNet \cite{yang2019xlnet} that proposed the permutation modeling to prevent the corruption of the input with masks, and RoBERTa \cite{liu2019roberta}, which can be seen as a BERT optimization that includes additional pre-training and hyperparameter improvements. 

We use the HuggingFace python library \cite{Wolf2019HuggingFacesTS}, which includes pre-trained models, in order to fine-tune each model to a classifier for the NLI task and a regressor for the QA task. We used the hyperparameters specified in the original paper for each model, to achieve an accuracy close to the ones reported for each task. 

Additionally, we include pre-Transformer baselines as a comparison reference. These models rely on the LSTM architecture \cite{hochreiter1997long} and are task-dependent. However, our analysis and discussion are mainly about experiments on Transformer-based models.

\section{NLI Task Description}
\label{NLI}
\subsection{Task}
The MultiNLI corpus \cite{N18-1101} is a crowd-sourced collection of 433k sentence pairs annotated with textual entailment information  from a broad range of genres. In this task, given a \textit{premise}, the model has to determine whether a \textit{hypothesis} is true (entailment), false (contradiction), or undetermined (neutral).

\subsection{Baselines}
As a baseline to evaluate stress test performance for this task, we chose the winner of RepEval 2017 Shared Task \cite{nangia-etal-2017-repeval}, which proposed a model of stacked BiLSTMs with residual connections \cite{nie-bansal-2017-shortcut}. Also, we used the baseline proposed in the original paper \cite{N18-1101} of the dataset, which consists of a standard BiLSTM.

\section{QA Task Description}
\label{QA}
\subsection{Task}
SQuAD, the Stanford Question Answering Dataset \cite{rajpurkar-etal-2016-squad} is a widely used Question Answering benchmark that consists of a collection of English Wikipedia paragraphs with more than 100k associated question-answer pairs generated via crowdsourcing. The task is designed in a way that the solution to each question is literally contained in the corresponding paragraph, so the task is to predict the answer text span in the corresponding passage.
We use SQuAD v1.1 instead of SQuAD v2.0 to allow comparability with previous work.

\subsection{Baselines}
To be consistent with previous work, we used BiDAF \cite{Seo2016BidirectionalAF} and Match-LSTM \cite{Wang2016MachineCU} as baselines to compare stress tests against Transformer-based models. BiDAF consists of embedding, attention and modeling layers with a BiLSTM, that outputs a vector with information of the context and the query, and finally an output layer with probabilities indicating where the answer starts and ends in the context text.      
In the case of Match-LSTM, the model is an architecture that remembers important word-level matching results to get better predictions of the answers.

\section{Experiments}
\label{experiments}

\subsection{NLI Task Evaluation}
Our experiments on the MultiNLI dataset closely follow the \newcite{naik18coling} procedure, which conducted a stress test evaluation of several models of the RepEval 2017 Shared Task. Below we describe each test set\footnote{We use the sets provided by the authors to avoid discrepancy during the procedure. \url{abhilasharavichander.github.io/NLI_StressTest}} used in this work  and  Table~\ref{table:nliexamples} shows some examples, however for further details of the sets construction we refer the readers to the work by \newcite{naik18coling}. 

\begin{table*}[h!]
\begin{tabular}{|l|l|l|}
\hline
\textbf{Test Set}                                                       & \textbf{Premise}                                                                                                                             & \textbf{Hypothesis}                                     \\ \hline
\textbf{\begin{tabular}[c]{@{}l@{}}Word \\ Overlap\end{tabular}}        & Then he ran.                                                                                                                                 & He ran like an athlete \textcolor{blue}{and true is true}.                \\ \hline
\textbf{\begin{tabular}[c]{@{}l@{}}Length \\ Mismatch\end{tabular}}     & \begin{tabular}[c]{@{}l@{}}Then he ran \textcolor{blue}{and true is true and true is true} \\ \textcolor{blue}{and true is true and true is true and true is true}.\end{tabular} & He ran like an athlete.                                 \\ \hline
\textbf{Negation}                                                       & Then he ran.                                                                                                                                 & He ran like an athlete \textcolor{blue}{and false is not true}.           \\ \hline
\textbf{\begin{tabular}[c]{@{}l@{}}Spelling \\ Errors\end{tabular}}     & Then he ran.                                                                                                                                 & He ran like an \textcolor{blue}{athleet}.                                 \\ \hline
\textbf{Antonymy}                                                       & The Joint Venture had justified itself by failure.                                                                                           & The Joint Venture had justified itself by \textcolor{blue}{success}.      \\ \hline
\textbf{\begin{tabular}[c]{@{}l@{}}Numerical \\ Reasoning\end{tabular}} & Adam spent 1/6 of his lifetime in adolescence.                                                                                               & Adam spent \textcolor{blue}{less than} 1/6 of his lifetime in adolescence. \\ \hline
\end{tabular}
\caption{Examples of stress tests for the NLI task.}
\label{table:nliexamples}
\end{table*}

\subsubsection{Distraction Test}
The distraction test explores the model robustness after a text with a clear ``True" value is added.
\begin{itemize}
  \item One way to evaluate this is by decreasing the lexical similarity between \textit{premise} and \textit{hypothesis}. On the one hand, the \textit{word overlap} set adds a tautology (``and true is true") at the end of each \textit{hypothesis} sentence. On the other hand, the \textit{length mismatch} set adds five times the same tautology to each \textit{premise}.
  \item We can also evaluate this by the inclusion of strong negations. The \textit{negation} set is quite similar to the previous ones, but in this case, the tautology added to the \textit{hypothesis} includes negation words (``and false is not true").
\end{itemize}

\subsubsection{Noise Test}
\label{nli-noise-test-specs}
This test verifies the model strength against noisy data, in terms of \textit{spelling errors}. It has two types of permutations on a word randomly selected from the \textit{hypothesis}: swap of adjacent characters within the word, and random substitution of a character next to it on the English keyboard. Note that only one substitution is performed for the entire sentence.

\subsubsection{Competence Test}
The competence test consists of two evaluation sets to measure the reasoning ability of the models.
\begin{itemize}
  \item Understanding of \textit{antonymy relationships}. This set includes sentences that result in contradiction simply by using an antonym in some adjectives or nouns. 
  \item \textit{Numerical reasoning} ability of a model. This evaluation includes statements of simple algebraic problems with solutions as \textit{premises}. The entailed, contradictory and neutral hypotheses were generated through the use of heuristic rules.
\end{itemize}

\begin{table*}[]
\begin{center}
\setlength\tabcolsep{1.71mm}
\begin{tabular}{|l|c|c|c|c|c|c|c|c|c|c|c|c|c|c|}
\cline{1-11}  \cline{13-15}
\multicolumn{1}{|c|}{\multirow{4}{*}{\textbf{Model}}} & \multicolumn{2}{c|}{\multirow{3}{*}{\textbf{\begin{tabular}[c]{@{}c@{}}Original\\ Dev\end{tabular}}}} & \multicolumn{6}{c|}{\textbf{Distraction Test}} & \multicolumn{2}{c|}{\textbf{Noise Test}} && \multicolumn{3}{c|}{\textbf{Competence Test}} \\ \cline{4-11}  \cline{13-15} 
\multicolumn{1}{|c|}{} & \multicolumn{2}{c|}{} & \multicolumn{2}{c|}{\textbf{\begin{tabular}[c]{@{}c@{}}Word \\ Overlap\end{tabular}}} & \multicolumn{2}{c|}{\textbf{Negation}} & \multicolumn{2}{c|}{\textbf{\begin{tabular}[c]{@{}c@{}}Length \\ Mismatch\end{tabular}}} & \multicolumn{2}{c|}{\textbf{\begin{tabular}[c]{@{}c@{}}Spelling \\ Error\end{tabular}}} && \multicolumn{2}{c|}{\textbf{Antonymy}} & \multirow{2}{*}{\textbf{\begin{tabular}[c]{@{}c@{}}Numerical\\ Reasoning\end{tabular}}} \\ \cline{2-11}  \cline{13-14}
\multicolumn{1}{|c|}{} & \textbf{M} & \textbf{MM} & \textbf{M} & \textbf{MM} & \textbf{M} & \textbf{MM} & \textbf{M} & \textbf{MM} & \textbf{M} & \textbf{MM} && \textbf{M} & \textbf{MM} & \\ 
\cline{1-11}  \cline{13-15}
\multirow{2}{*}{RoBERTa} & \multirow{2}{*}{\textbf{90.0}} & \multirow{2}{*}{\textbf{89.7}} & 64.3 & 62.3 & 59.0 & 58.5 & \textbf{87.5} & \textbf{88.2} & \textbf{85.3} & \textbf{85.7} && \multirow{2}{*}{63.9} & \multirow{2}{*}{59.2} & \multirow{2}{*}{64.9} \\
 & & & {\color{mygray}\footnotesize[28.5]} & {\color{mygray}\footnotesize[30.5]} & {\color{mygray}\footnotesize[34.4]} & {\color{mygray}\footnotesize[34.8]} & {\color{mygray}\footnotesize[2.8]} & {\color{mygray}\footnotesize[\textbf{1.7}]} & {\color{mygray}\footnotesize[\textbf{5.2}]} & {\color{mygray}\footnotesize[\textbf{4.5}]} &&  &  &  \\
\cline{1-11}  \cline{13-15}
\multirow{2}{*}{XLNet} & \multirow{2}{*}{89.2} & \multirow{2}{*}{89.1} & \textbf{71.0} & \textbf{68.9} & \textbf{60.0} & \textbf{59.5} & 87.2 & 87.5 & 83.5 & 83.7 && \multirow{2}{*}{74.7} & \multirow{2}{*}{70.9} & \multirow{2}{*}{63.9} \\ 
 & & & {\color{mygray}\footnotesize[20.4]} & {\color{mygray}\footnotesize[22.7]} & {\color{mygray}\footnotesize[32.7]} & {\color{mygray}\footnotesize[33.2]} & {\color{mygray}\footnotesize[\textbf{1.9}]} & {\color{mygray}\footnotesize[1.8]} & {\color{mygray}\footnotesize[6.4]} & {\color{mygray}\footnotesize[6.1]} && & & \\
\cline{1-11}  \cline{13-15}
\multirow{2}{*}{BERT} & \multirow{2}{*}{86.0} & \multirow{2}{*}{86.1} & 61.2 & 56.8 & 57.3 & 57.6 & 83.7 & 84.6 & 79.5 & 79.8 && \multirow{2}{*}{64.6} & \multirow{2}{*}{59.2} & \multirow{2}{*}{56.8} \\ 
 & & & {\color{mygray}\footnotesize[28.8]} & {\color{mygray}\footnotesize[34.0]} & {\color{mygray}\footnotesize[33.4]} & {\color{mygray}\footnotesize[33.1]} & {\color{mygray}\footnotesize[2.7]} & {\color{mygray}\footnotesize[\textbf{1.7}]} & {\color{mygray}\footnotesize[7.6]} & {\color{mygray}\footnotesize[7.3]} && & & \\
\cline{1-11}  \cline{13-15}
\multirow{2}{*}{S-BiLSTM} & \multirow{2}{*}{74.2} & \multirow{2}{*}{74.8} & 47.2 & 47.1 & 39.5 & 40.0 & 48.2 & 47.3 & 51.1 & 49.8 && \multirow{2}{*}{15.1} & \multirow{2}{*}{19.3} & \multirow{2}{*}{21.2} \\ 
 & & & {\color{mygray}\footnotesize[36.4]} & {\color{mygray}\footnotesize[37.0]} & {\color{mygray}\footnotesize[46.8]} & {\color{mygray}\footnotesize[46.5]} & {\color{mygray}\footnotesize[35.0]} & {\color{mygray}\footnotesize[36.8]} & {\color{mygray}\footnotesize[31.1]} & {\color{mygray}\footnotesize[33.4]} && & & \\
\cline{1-11}  \cline{13-15}
\multirow{2}{*}{BiLSTM} & \multirow{2}{*}{70.2} & \multirow{2}{*}{70.8} & 57.0 & 58.5 & 51.4 & 51.9 & 49.7 & 51.2 & 65.0 & 65.1 && \multirow{2}{*}{13.2} & \multirow{2}{*}{9.8} & \multirow{2}{*}{31.3} \\ 
 & & & {\color{mygray}\footnotesize[\textbf{18.8}]} & {\color{mygray}\footnotesize[\textbf{17.4}]} & {\color{mygray}\footnotesize[\textbf{26.8}]} & {\color{mygray}\footnotesize[\textbf{26.7}]} & {\color{mygray}\footnotesize[29.2]} & {\color{mygray}\footnotesize[27.7]} & {\color{mygray}\footnotesize[7.4]} & {\color{mygray}\footnotesize[8.1]} && & & \\
\cline{1-11}  \cline{13-15}
\end{tabular}
\caption{Classification accuracy (\%) of Transformer-based models and baselines. Both genre-matched (M) and mismatched (MM) sets were evaluated. Values in brackets represent the percentage of reduction with respect to the original dev set.}
\label{table:nlitests}
\end{center}
\end{table*}

\subsection{NLI Task Results}

Table~\ref{table:nlitests} shows the results of the performed tests. It can be seen that all models decrease their accuracy in all evaluations. However, Transformer-based models show more robustness in some tests. The analysis of the results of the models in each stress test is shown on the following sections.

\begin{figure}[]
\begin{center}
\includegraphics[width=0.46\textwidth]{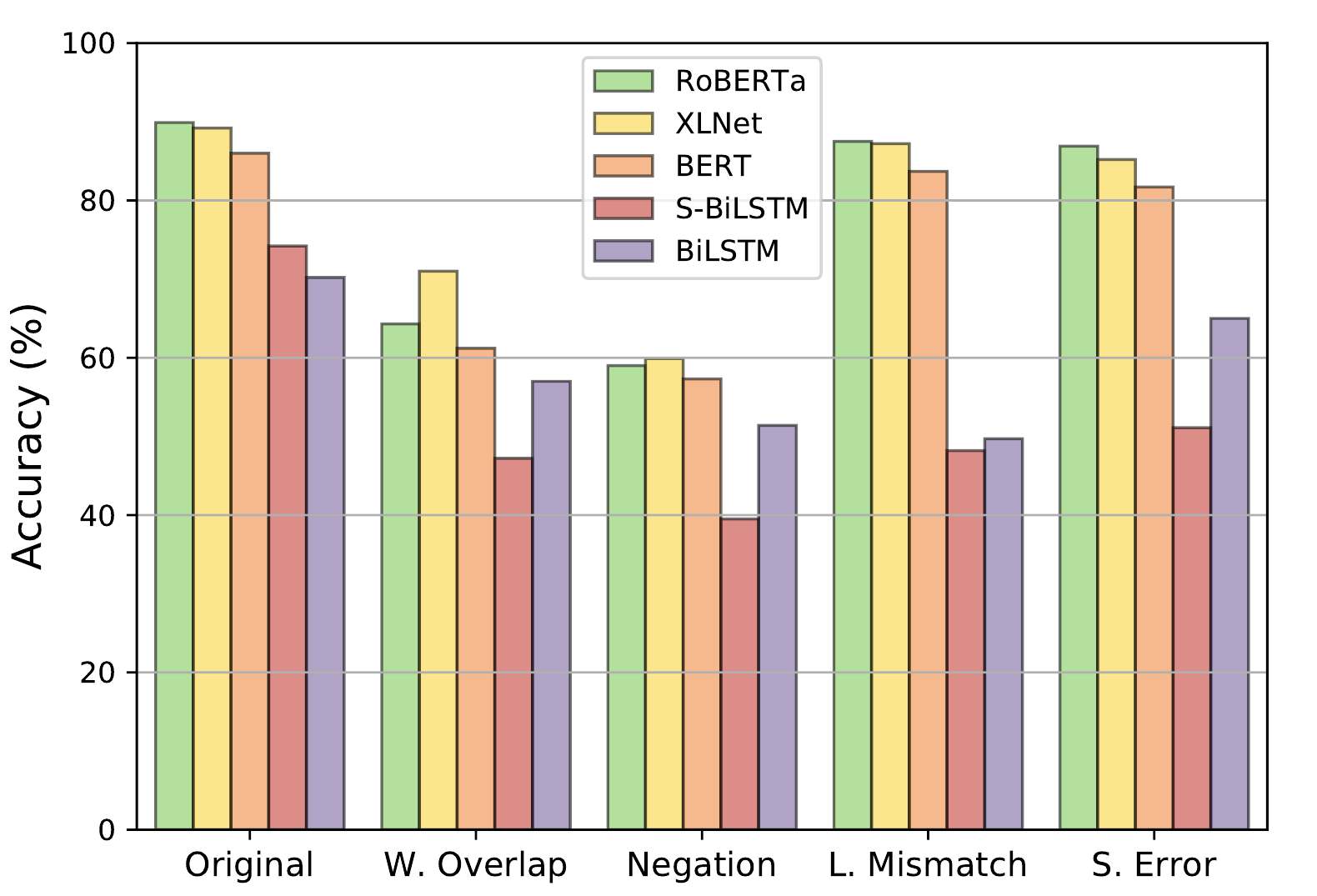} 
%\vspace{-10mm}
\caption{Accuracy results in the development set and adversarial sets: word overlap, negation, length mismatch and spelling error. Only matched partition is shown.}
\label{fig:nlitests}
\end{center}
\vspace{-3mm}
\end{figure}

\subsubsection{Models Performance on Distraction Test}
\label{distracion_sec}
Figure~\ref{fig:nlitests} shows a bar graph of the ``matched" partition of the evaluation sets on the different types of distraction tests. As mentioned in a previous section, the distraction tests allow us to check the robustness in two different ways.

On the one hand, the effect of introducing \textit{negation} words drops the models performance below 60\% of accuracy, close to the baselines. We checked the model predictions on the negation test v/s the development set and we found that BERT and XLNet obtained 93\% and  91\% of E-N (entailment predicted as neutral) error respectively. In contrast, RoBERTa obtained 85\% of N-E error (neutral predicted as entailment). This could occur due to the introduction of extra negation words (``false" and ``not").

On the other hand, the decrease of lexical similarity by \textit{word overlap} and \textit{length mismatch} evaluation shows:

\begin{itemize}
    \item In the first case (\textit{word overlap} set), the Transformer-based models reach around 60\% accuracy, which is approximately 20\% less than in the development set. We found a similar behavior with the previous set (\textit{negation}), where BERT and XLNet obtained 83\% and 61\% of E-N error respectively. It also stands out that RoBERTa achieved 89\% of N-E error.
    \item In the second case (\textit{length mismatch}), the models performed better than expected, because they reached almost the same accuracy as in the development set. We hypothesize that these results may be due to the \textit{length mismatch} set modifying the \textit{premise} sentence instead of the \textit{hypothesis} as in the negation of the \textit{word overlap} sets, which suggests that in order to answer, the model is paying more attention to that sentence.
\end{itemize}

\begin{figure}[]
\begin{center}
\includegraphics[width=0.49\textwidth]{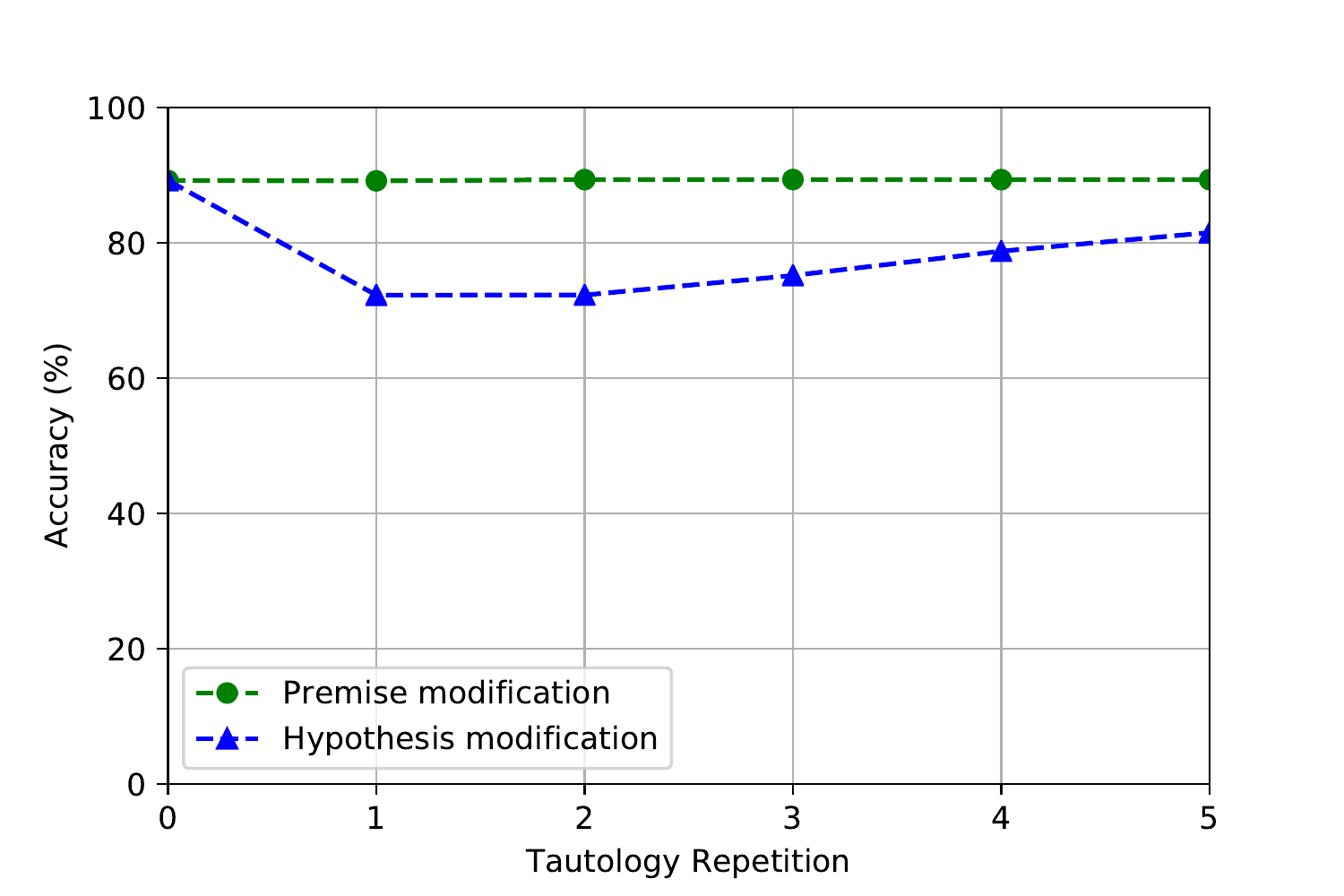} 
%\vspace{-10mm}
\caption{Accuracy (\%) of XLNet after the addition of different number of tautologies in hypothesis or premises.}
\label{fig:nlitaut}
\end{center}
\vspace{-3mm}
\end{figure}

To verify the results on the \textit{length mismatch} set, we extended the evaluation by testing the addition of the tautology ``and true is true" in the \textit{hypothesis} or in the \textit{premises} $N$ times (where $N=1..5$). Figure~\ref{fig:nlitaut} shows the performance of XLNet in these tests, likewise we observed similar behavior on the other models. We noticed that the inclusion of the distractions to the premise sentence does not affect the model performance. However, when we add the tautology a single time (which is equivalent to the \textit{word overlap} test) to the \textit{hypothesis} sentence, the performance drops about 20\%, and the more repetitions we add, the more accuracy increases, almost reaching the same performance obtained in the development set. We also checked the attention weights, and did not identify anomalous behavior.

The unexpected result in accuracy indicates that the lexical similarity is not a strong enough signal to generate distraction in this type of model, as they can discern the tautologies anyway. Moreover, the model seems to pay more attention to the \textit{hypothesis} sentence in order to respond, without discarding the \textit{premise}. However, the distraction evaluation indicates that these Transformer-based models are fragile to adversarial attacks that include strong negation words.

\subsubsection{Models Performance on Noise Test}
The noise test with the \textit{spelling error} set exhibits that Transformer-based models perform very well. They only lose between 2 to 5 percentage points in accuracy with respect to the \textit{development} set. The results suggest that the multi-head self-attention mechanism of these models is very effective at recovering the global information from the corrupted sentence.

However, the adversarial attacks of this set only modify one word of the \textit{hypothesis}. This explains why there is no sudden drop in performance in models, even for the BiLSTM-based models.

\subsubsection{Models Performance on Competence Test}
As we supposed, Transformer-based models work quite well in this evaluation task. In the case of the \textit{antonymy} test, the models exceeded baselines by approximately 50 percentage points in accuracy. This is probably because Transformers were pre-trained on a diverse and big corpus, allowing them to adequately represent the majority of the words of the dictionary. XLNet and BERT were trained with BookCorpus and Wikipedia, so we expected better accuracy of RoBERTa which used additional data. However, XLNet outperformed others by at least 10 percentage points, suggesting that permutation modeling could help capture antonymy relationships better.

Furthermore, the results on the \textit{numerical reasoning} evaluation show a lower performance for all models. In this task, XLNet and RoBERTa have similar accuracy but have different behavior. On the one hand, XLNet specialized in classifying ``entailment" samples, achieving 90\% in that class. On the other hand, RoBERTa specialized in ``neutral" category, obtaining 89\% of correct answers. In both cases, the remaining classes achieved less than 74\% of accuracy (the model finds it hard to distinguish between those classes). These results indicate that Transformer-based models trained in the NLI task have serious difficulties in numerical reasoning and that they take different strategies to solve the task.

For both evaluations, we also explored the attention weights via the \textit{BertViz} library \cite{vig2019Transformervis}. Appendix B shows a brief analysis of some specific cases on all the mentioned Transformer-based models.

\subsubsection{Annotation Artifacts Exploitation Test}
\label{annotation_sec}
\newcite{gururangan-etal-2018-annotation} found that MultiNLI dataset has annotation artifacts. It means that crowd workers who participated in the creation of the data, adopted heuristics to generate the \textit{hypothesis} in an easy and fast way. For instance, they usually use some keywords such as ``not", ``never", etc. to create negation sentences.

To evaluate if Transformer-based models leverage the artifacts, we tested the models by removing the \textit{premise} sentence in the development set. In other words, the models are unaware of the \textit{premises} of the dataset.

Table~\ref{table:nliartifacts} shows the results of this experiment. It is possible to see that Transformer-based models perform similar to the majority class\footnote{The majority class is used as a baseline of random guessing.}, which denotes an unbiased guess of the models. In contrast, BiLSTM-based models show significant proportion of correctly classified samples without even looking at the \textit{premise} (which is an undesirable behavior). This result demonstrates that Transformer-based models are in fact learning to take into account and relate the two sentences of the NLI task in order to choose the correct answer, which is consistent with the findings in Section~\ref{distracion_sec}.

\begin{table}[]
\begin{center}
\begin{tabular}{l|c|c|}
\cline{2-3}
                                     & \multicolumn{1}{l|}{\textbf{Matched}} & \multicolumn{1}{l|}{\textbf{Mismatched}} \\ \hline
\multicolumn{1}{|l|}{Majority Class} & 35.4                                  & 35.2                                     \\ \hline
\multicolumn{1}{|l|}{RoBERTa}        & 35.2                                  & 35.8                                     \\ \hline
\multicolumn{1}{|l|}{XLNet}          & 35.4                                  & 35.8                                     \\ \hline
\multicolumn{1}{|l|}{BERT}           & 35.5                                  & 35.7                                     \\ \hline
\multicolumn{1}{|l|}{S-BiLSTM}       & 45.2                                  &  45.4                                     \\ \hline
\multicolumn{1}{|l|}{BiLSTM}         & 37.4                                  &  38.3                                    \\ \hline
\end{tabular}
\end{center}
\caption{Performance (\%) of premise-unaware text models on MultiNLI development set. Greater accuracy means more exploitation of artifacts, thus smaller numbers mean the models performed best.}
\label{table:nliartifacts}
\end{table}

\subsection{QA Task Evaluation}
One of our test scenarios was taken from \newcite{jia-liang-2017-adversarial}, which intentionally adds a new adversarial sentence at the end of SQuAD passages of the development set. These sentences are especially designed (via different strategies) to act as a decoy to confuse the model.
The other test scenario is inspired on \newcite{belinkov2018synthetic}. Although originally proposed for a different task, we replicated the 5 types of noise proposed by the authors, and applied them on the development set of SQuAD. 

\subsubsection{Adversarial Sentence Tests}
In \newcite{jia-liang-2017-adversarial}, the authors proposed 4 strategies to create a sentence especially designed to confuse models by pretending to be the correct answer to a specific question, although they are unrelated with the question. This adversarial sentence is concatenated to the corresponding paragraph provided at test time. The 4 strategies proposed were:
\begin{itemize}
    \item \textbf{AddOneSent}: Adjectives and nouns of the question are replaced by antonyms. Named entities and numbers are replaced by their nearest word in GloVe \cite{pennington-etal-2014-glove}. This modified question is then turned into declarative form (using a set of manually defined rules) and a fake answer of the same type as the original answer is inserted. Finally the sentence is manually checked and fixed via crowdsourcing. %\vladimir{dejamos is o mejor was?} \andres{is quizas pq es un experimento estandarizado} \charlie{OK.}
    \item \textbf{AddSent}: Identical to \textit{AddOneSent} but generating multiple candidate sentences (adversaries) and keeping only the one that induces the biggest error when tested on a specific model.
    \item \textbf{AddAny}: The adversarial sentence is generated by sampling random words and successively replacing them by elements from a sampled set of $20$ words each time. Words are selected from this set  by using a criterion that tries to minimize the confidence of the model on the correct answer. The 20-word set is sampled from a list of common words plus the words from the question. This process is repeated iteratively 6 times for each adversarial phrase. 
    \item \textbf{AddCommon}: Identical to \textit{AddAny}, but in this case the 20-word set is sampled from the list of common words directly.
\end{itemize}

\begin{figure}[t]
    \small
    \begin{framed}
    \footnotesize
      \textbf{Article:} Super Bowl 50
    
      \textbf{Context:}
      {Peyton Manning became the first quarterback ever to lead two different teams to multiple Super Bowls. He is also the oldest quarterback ever to play in a Super Bowl at age 39. The past record was held by John Elway, who led the Broncos to victory in Super Bowl XXXIII at age 38 and is currently Denver's Executive Vice President of Football Operations and General Manager.
      \textcolor{blue}{\textit{Quarterback Jeff Dean had jersey number 37 in Champ Bowl XXXIV.}}}
    
      \textbf{Question:} {What is the name of the quarterback who was 38 in Super Bowl XXXIII?}
    
      \textbf{Original prediction:} {John Elway}
    
      \textbf{Prediction after adversarial phrase is added:} \textcolor{red}{Jeff Dean}
    
    \end{framed}
    \caption{
      An example of an \textit{AddOneSent} adversarial sample. This example was taken from \newcite{jia-liang-2017-adversarial}. In this case we can see that the model correctly answered the original question, but after the inclusion of the adversarial sentence (in \textcolor{blue}{\textit{italic blue}}), the model fails (answer in \textcolor{red}{red}).
    }
    \label{fig:addonesent-example}
\end{figure}

\subsubsection{Noise Tests}
Although originally proposed for a different task, we replicated the 5 types of noise introduced by \newcite{belinkov2018synthetic}. In each experiment, a specific noise type was applied to each word in the passage of SQuAD's development set. The question was kept unchanged, and the answers were adapted to preserve consistency with the modified passage. 
In contrast to the noise tests performed in the NLI setting (Section~\ref{nli-noise-test-specs}), the scenario tested here is significantly more aggressive because it introduces noise to every word in the reference text.

The 5 noise types tested are:
\begin{itemize}
    \item \textbf{Natural Noise}: Words are replaced by real typing errors of people. To automate this, we used a collection of word corrections performed by people in web platforms that keep track of edits history \cite{max-wisniewski-2010-mining,zesch-2012-measuring,wisniewski-merlin-2013,11234/1-2143}.
    \item \textbf{Swap Noise}: For each word in the text, one random pair of consecutive characters is swapped (e.g. $expression \rightarrow exrpession$).
    \item \textbf{Middle Random Noise}: For each word in the text, all characters are shuffled, except for the first and last characters. (e.g. $expression \rightarrow esroxiespn$).
    \item \textbf{Fully Random Noise}: For each word in the text, all characters are shuffled (e.g. $expression \rightarrow rsnixpoees$).
    \item \textbf{Keyboard Typo Noise}: For each word in the text, one character is replaced by an adjacent character in traditional English keyboards (e.g. $expression \rightarrow exprwssion$).
\end{itemize}

\subsection{QA Task Results}

Similarly to the observations for the NLI experiments, for QA it is clear that the performance of all models is affected by the stress tests, with Transformer-based models being the most robust in all the cases analyzed. Detailed results can be found in Table~\ref{table:squadtestssummary}.

\begin{table*}[]
\setlength\tabcolsep{1.02mm}
\begin{tabular}{|l|c|c|c|c|c|c|c|c|c|c|}
\hline
\multicolumn{1}{|c|}{\multirow{3}{*}{\textbf{Model}}} & \multirow{3}{*}{\begin{tabular}[c]{@{}c@{}}\textbf{Original} \\ \textbf{Dev}\end{tabular}} & \multicolumn{4}{c|}{\textbf{Concatenative Adversaries}} & \multicolumn{5}{c|}{\textbf{Noise Adversaries}} \\
\cline{3-11}
{ }  & & \begin{tabular}[c]{@{}c@{}}\textbf{AddOne-} \\ \textbf{Sent}\end{tabular} & \textbf{AddSent} & \textbf{AddAny} & \begin{tabular}[c]{@{}c@{}}\textbf{Add-} \\ \textbf{Common}\end{tabular} & { } \textbf{Swap} { } & \begin{tabular}[c]{@{}c@{}}\textbf{Middle} \\ \textbf{Random} \end{tabular} & \begin{tabular}[c]{@{}c@{}}\textbf{Fully} \\ \textbf{Random}\end{tabular} & \begin{tabular}[c]{@{}c@{}}\textbf{Keyboa-} \\ \textbf{rd Typo}\end{tabular} & \begin{tabular}[c]{@{}c@{}}\textbf{Natural}\end{tabular} \\ \hline
\multirow{2}{*}{RoBERTa}     & \multirow{2}{*}{\textbf{85.8}} & \textbf{70.3} & 61.5 & 77.3 & \textbf{84.3} & \textbf{46.1} & \textbf{32.2} & 3.3 & \textbf{30.4} & 54.9 \\
 & & {\color{mygray}\footnotesize[\textbf{18.1}]}& {\color{mygray}\footnotesize[28.3]}& {\color{mygray}\footnotesize[9.9]}& {\color{mygray}\footnotesize[\textbf{1.7}]}& {\color{mygray}\footnotesize[\textbf{46.3}]}& {\color{mygray}\footnotesize[\textbf{62.5}]}& {\color{mygray}\footnotesize[96.2]}& {\color{mygray}\footnotesize[\textbf{64.6}]}& {\color{mygray}\footnotesize[36.0]} \\
\hline
\multirow{2}{*}{XLNet}       & \multirow{2}{*}{85.2} & 67.7 & \textbf{61.6} & \textbf{78.8} & 83.0 & 43.0 & 31.9 & 4.4 & 27.2 & \textbf{57.4} \\
 & & {\color{mygray}\footnotesize[20.5]}& {\color{mygray}\footnotesize[\textbf{27.7}]}& {\color{mygray}\footnotesize[\textbf{7.5}]}& {\color{mygray}\footnotesize[2.6]}& {\color{mygray}\footnotesize[49.5]}& {\color{mygray}\footnotesize[62.6]}& {\color{mygray}\footnotesize[94.8]}& {\color{mygray}\footnotesize[68.1]}& {\color{mygray}\footnotesize[\textbf{32.6}]} \\
\hline
\multirow{2}{*}{BERT}        & \multirow{2}{*}{82.5} & 64.6 & 55.9 & 71.4 & 81.1 & 33.8 & 28.6 & \textbf{5.5} & 23.1 & 47.7 \\
 & & {\color{mygray}\footnotesize[21.7]}& {\color{mygray}\footnotesize[32.2]}& {\color{mygray}\footnotesize[13.5]}& {\color{mygray}\footnotesize[\textbf{1.7}]}& {\color{mygray}\footnotesize[59.0]}& {\color{mygray}\footnotesize[65.3]}& {\color{mygray}\footnotesize[\textbf{93.3}]}& {\color{mygray}\footnotesize[72.0]}& {\color{mygray}\footnotesize[42.2]} \\
\hline
\multirow{2}{*}{Match-LSTM}  & \multirow{2}{*}{60.8} & 30.0 & 24.8 & 35.7 & 52.5 & 17.8 & 20.2 & 4.1 & { }9.4 & 19.7 \\
 & & {\color{mygray}\footnotesize[50.7]}& {\color{mygray}\footnotesize[59.2]}& {\color{mygray}\footnotesize[41.3]}& {\color{mygray}\footnotesize[13.7]}& {\color{mygray}\footnotesize[70.7]}& {\color{mygray}\footnotesize[66.8]}& {\color{mygray}\footnotesize[\textbf{93.3}]}& {\color{mygray}\footnotesize[84.5]}& {\color{mygray}\footnotesize[67.6]} \\
\hline
\end{tabular}
\caption{Exact match (\%) of Transformer-based models and baselines on the SQuAD v1.1 dev set. AddAny and AddCommon report the worst accuracy after running against all the alternative adversarial datasets of that specific type. For fair comparison, experiments on the adversaries generated for the model itself are excluded in those two specific cases. Values in brackets represent the percentage of reduction with respect to the original dev set.}
\label{table:squadtestssummary}
\vspace{-6mm}
\end{table*}

\subsubsection{Results on Adversarial Sentence Tests}
\label{results-jia-liang-adversaries}

\begin{figure}[]
\begin{center}
\includegraphics[width=0.5\textwidth]{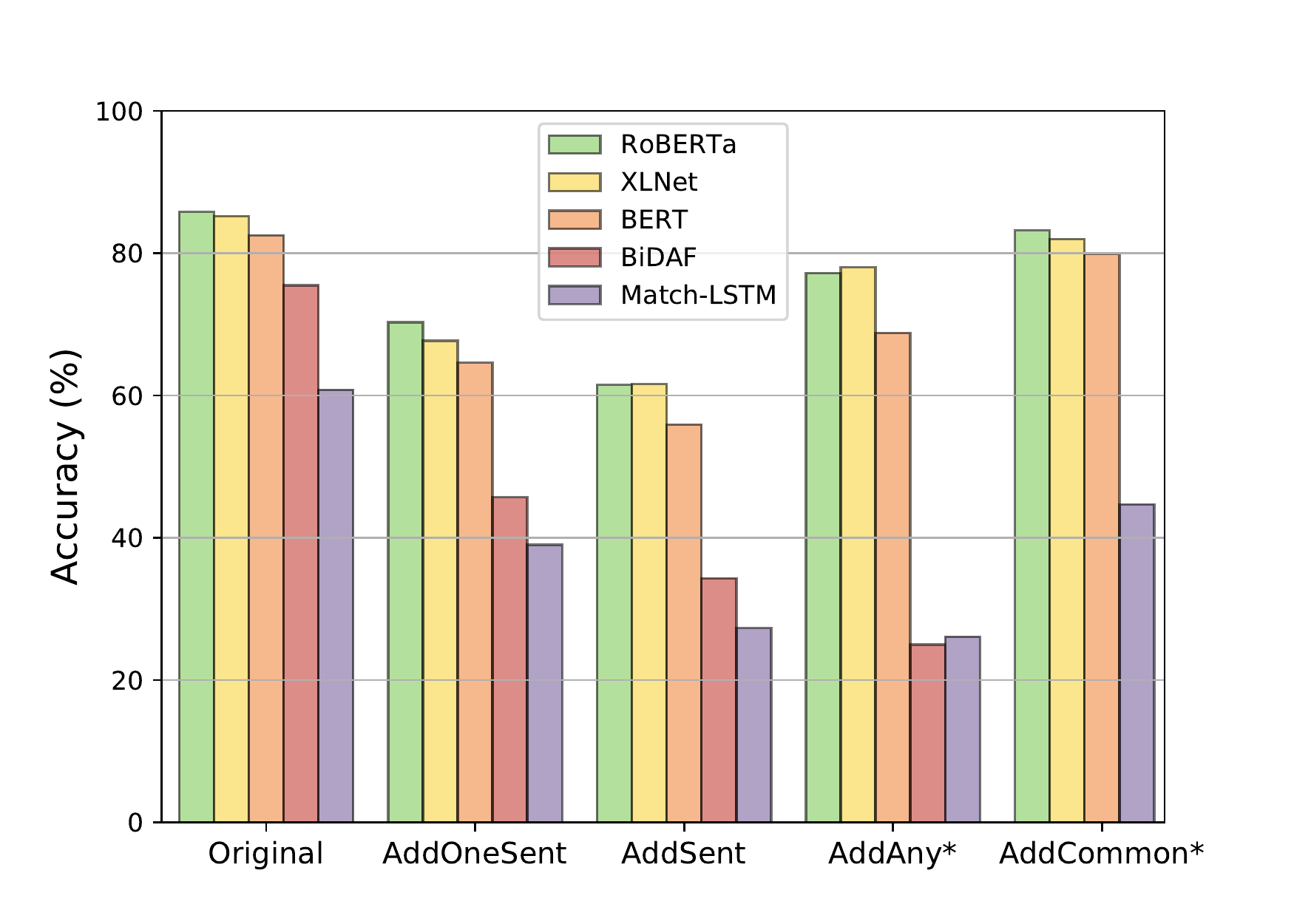}
\vspace{-6mm}
\caption{Accuracy results in the adversarial sets proposed by \newcite{jia-liang-2017-adversarial}. AddAny* and AddCommon* report the worst accuracy after running against all the alternative adversarial datasets of that specific type. For fair comparison, experiments on the adversaries generated for the model itself are excluded in those two specific cases.}
\label{fig:qa_jia_adversaries}
\end{center}
\vspace{-3mm}
\end{figure}

Figure~\ref{fig:qa_jia_adversaries} shows a bar graph that compares the accuracy of the tested models under the different adversarial strategies.

When we analyze the results of the \textit{AddOneSent} experiments, we notice an accuracy reduction between $18.1\%$ and $21.7\%$ for the Transformer-based models, and greater than $39.5\%$ for non-Transformer models. In spite of showing greater robustness in comparison with their counterpart, Transformer-based models still suffer from a significant impact on performance, which elucidates a clear opportunity for future improvements on these kind of models.
The same phenomenon is observed for \textit{AddSent} adversaries, but more pronounced (as expected, since \textit{AddSent} tests the worst case for each candidate question). We see accuracy reductions ranging from $27.7\%$ and $32.2\%$ for Transformer-based models, and greater than $54.6\%$ for non-Transformer models.

We notice that as the model is more powerful in the main task (accuracy in the unmodified SQuAD v1.1 development set), it also achieves greater robustness. This conclusion is hopeful because other works have asserted that more powerful models could justify their performance on their higher memorization capabilities \cite{45820}. These experiments, in contrast, indicate that the models are improving their reading capabilities in a balanced fashion.

Interestingly, \textit{AddAny} and \textit{AddCommon} adversaries show that those strategies are very model-specific, as evidenced by the fact that Transformer-based models only reduce their accuracy in small degree when tested against adversaries where other architectures failed. 
These results are relevant because, as reported by \newcite{jia-liang-2017-adversarial}, those adversaries (and especially \textit{AddAny}) turned to be very effective when trying to mislead the models that they were targeting. 
This cross-check between different model's adversaries for \textit{AddAny} is consistent with the results reported by \newcite{jia-liang-2017-adversarial}, although in the case of Transformer-based models, the before-mentioned behavior is even more pronounced. For the case of \textit{AddCommon}, in the other hand, this tests were not reported in previous work nor analyzed by the authors that proposed these adversaries, thus this finding is especially relevant.

Further details on the results of every experiment performed can be found in Appendix A. Also in Appendix C we perform a more qualitative analysis of the attention matrices that these models produce during inference.

\subsubsection{Results on Noise Tests}
\label{noise-squad-results-section}
\begin{figure}[]
\begin{center}
\includegraphics[width=0.52\textwidth]{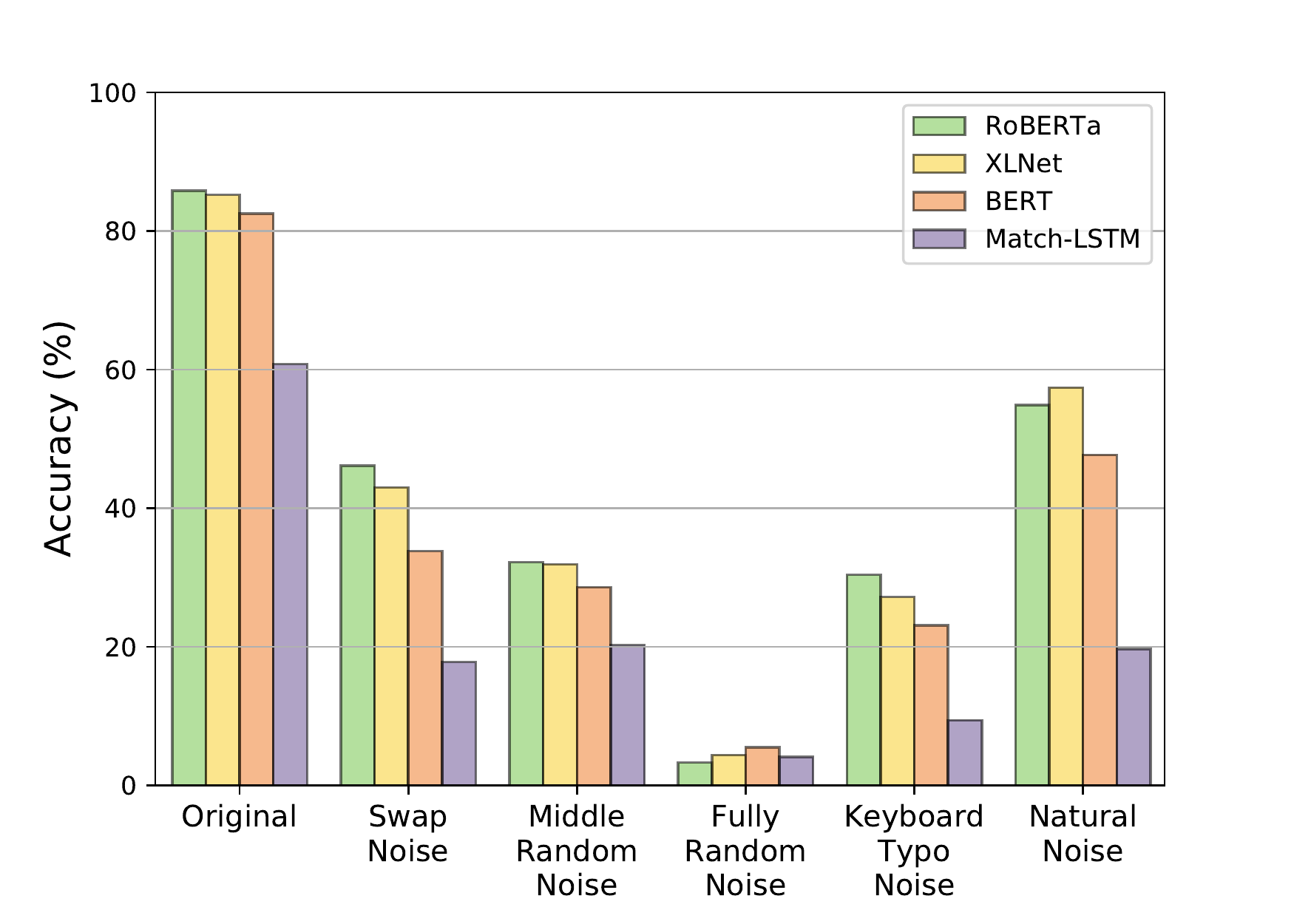}
\vspace{-6mm}
\caption{Accuracy results in SQuAD when the models are exposed to noise tests. It is clear that all noise types heavily affect the performance of all the models. Further comparative analysis (Section~\ref{noise-squad-results-section}) show some interesting and unexpected findings in these results.}
\label{fig:qa_noise_graph}
\end{center}
\vspace{-3mm}
\end{figure}

As shown in Figure~\ref{fig:qa_noise_graph}, all five types of noise have a significant negative impact on accuracy on all the tested models. The accuracy reduction is more prominent than on Adversarial Sentence tests (Section~\ref{results-jia-liang-adversaries}) due to the aggressiveness of the strategies tested here.

\textit{Swap Noise} has a significant impact on accuracy, between $46.3\%$ and $59.0\%$ (for the Transformer-based models) and of $70.1\%$ for Match-LSTM, although only a single pair of characters per word are altered. Performance is only slightly better than when using \textit{Middle Random Noise} (and in that scenario, all the characters are shuffled, except for the first and last ones). We hypothesize that this is due to the fact that by introducing this change, the resulting tokenization differ significantly from the original ones and are also very different from the ones seen in training or fine-tuning, and thus the model is not prepared to answer accurately.

\begin{figure}[t]
    \small
    \begin{framed}
    \footnotesize
      \textbf{Article:} Genghis Khan
    
      \textbf{Context:}
      {(...) Maluqi, a tsuretd lteneitnau, was given cmmnoad of the Monogl focres angisat the Jin dytasny whlie Gneghis Kahn was ftgniihg in Ctneral Aais, and Stbuaui and Jbee were aeolwld to prusue the Great Raid itno the Cuaucsas and Kaiven Rus', an idea tehy had peestrned to the Kaaghn on tehir own ieivtnitia. Whlie grnniatg his gneaelrs a gerat dael of amotonuy in mkiang canommd diesscion, Gnhgeis Kahn also epecxted uvwannrieg layolty from them.}
      
      \textbf{Question:} {Who was delegated command of the Mongol forces against the Jin dynasty?}
    
      \textbf{Answer:} {Maluqi}
    \end{framed}
    \vspace{-5mm}
    \caption{
      A QA adversarial example after the introduction of \textit{Middle Random} noise. Note that only the context (and the answer, accordingly) is modified, but not the question.
    }
    \label{fig:middle-random-example}
    % \vspace{-5mm}
\end{figure}

Note also that, in absolute terms, under \textit{Middle Random} noise, the model is still able to correctly answer one in four questions, even though the text is severely transformed (example in Figure~\ref{fig:middle-random-example}).

Another unexpected pattern that these tests showed is the fact that for Transformer-based models, the \textit{Keyboard Typo} noise is more challenging to deal with than \textit{Swap Noise}. This finding is especially intriguing because \textit{Keyboard Typo} noise corrupts only one character for each word, and \textit{Swap Noise} corrupts two. For this reason, this result is opposed to what we expected and reveals that \textit{swapping} operations affect these models less than \textit{replacement} operations. This effect may be caused by the fact that the tokenized representation of words with \textit{swapped} characters might be closer to the original one (in the embedding space of each model), or maybe it is because this kind of noise might be more frequent in real misspellings than keyboard typos, so the models were more exposed to this kind of noise during pre-training. Further study is required to find out which phenomenon is the dominant one in this case, but this analysis is out of the scope of this work.

Similarly to what was reported in \newcite{belinkov2018synthetic}, \textit{Natural Noise} is significantly more straightforward to overcome than the other four tested noise types, even considering that in the dataset we built for \textit{Natural Noise}, we forcefully replaced every word by a noisy version of it (when real typing errors were available).
It is natural to think that in real scenarios, misspelled words will appear at a much lower rate than in this test. Thus this result can be seen as a kind of lower-bound estimator for performance on \textit{Natural Noise} in real scenarios. When we compare the result of the \textit{Natural Noise} experiments with those of the \textit{Swap Noise} experiments, we hypothesize that the gap in favor to \textit{Natural Noise} is because, during the pre-training phase, the model observed this type of noise (in real occurrences) and was, therefore, able to learn useful representations both for well-written words and for versions with common misspellings.

\section{Related Work and Discussion}
\label{RW}
Prior work \cite{smith2012adversarial} discusses the importance of evaluation frameworks that allow characterizing model successes and failures. During previous years, several approaches to test NLP models have been proposed on various tasks, showing that most of the time, predictions are memorized without really understanding the real meaning of utterances \cite{45820}. 

Early research demonstrated that NLP models are fragile to input perturbations. Some attempts at performing stress tests on machine translation systems demonstrated that by adding small perturbations on the input text, the general performance of language models could be profoundly affected \cite{belinkov2018synthetic,ribeiro-etal-2018-semantically,ebrahimi-etal-2018-adversarial}. In the same line, the inspiring work of \newcite{jia-liang-2017-adversarial} proposed an evaluation procedure for language models using the SQuAD dataset. They used SQuAD samples, concatenating adversarial sentences at the end of the paragraph that contains the answer, and showed that 14 open-source models failed when these changes are introduced.

Other relevant findings reveal that models take advantage of lexical cues of the dataset, allowing them to solve the problem falsely. \newcite{gururangan-etal-2018-annotation} observed that some NLI datasets have annotation artifacts that models exploit to predict the answer without even seeing the rest of the sentence. The same problem was found in the Visual Question Answering (VQA) field. \newcite{agrawal-etal-2016-analyzing} analyzed the behavior of three models based on CNN, LSTM, and attention mechanism by adding adversaries only to the caption of the image, obtaining that most of the times models were paying attention to the text and not the image at inference time. 

The success of language models based on the Transformer architecture in tasks such as machine translation \cite{vaswani2017attention,vaswani2018tensor2tensor}, text summarization \cite{kroening2008loop}, reading comprehension \cite{dehghani2018universal}, among others, motivated new research. Recent works have performed adversarial testing of BERT in Argument Reasoning Comprehension Task \cite{niven2019probing}. They have shown that tested against adversaries, BERT outperforms BiLSTM and Bag of Vectors baselines, but still has trouble with logic understanding.
Furthermore, \newcite{jin2019bert} showed that BERT is the language model that best performs under adversary attacks when compared to CNN and LSTM in terms of success rate and perturbation rate, preservation of semantic content, and efficiency for text classification tasks. \newcite{hsieh-etal-2019-robustness} also studied BERT and compared it with recurrent architectures, inspecting the attention matrices of the models and proposing an algorithm to generate adversaries focusing on distracting models but not humans.

Although there is considerable progress in this area, it can be seen that this article differentiates from previous works by systematically evaluating adversaries, artifacts and various severe stress conditions on the state-of-the-art language models based on Transformer (BERT and the models that came after it), in order to verify their language comprehension capabilities and generalization power. 

As a final thought, to use these models in real-world applications, the reader must take the conclusions exposed in this work carefully, as some of the noise types and adversaries are far more aggressive than what can be expected in real scenarios. Rather than defining a realistic test scenario, the purpose of this work was to study these models robustness under severe stress conditions to elucidate their strengths and weaknesses, and in some cases quantify an upper bound in the impact that noise or misleading information can have in them. Additionally, some of the adversarial datasets of this work could be used to improve the robustness of the models through an adversarial training process, as they can be seen as an exaggerated version of common typographical errors made by humans.

\section{Conclusion}
\label{conclusions}
We conducted a stress test evaluation for Transformer-based language models in NLI and QA tasks. In general, our experiments indicate that applying stress tests influenced the performance of all models, but as expected, more recent models such as XLNet and RoBERTa are more robust, showing a better response to this evaluation. 

In the NLI task, we verified that the distraction test significantly reduces the performance of all models, especially in the negation test. However, tests on noise examples show that models are somewhat robust, possibly because they were pre-trained in a huge corpus that may have had natural noise. Due to the same reason, the models show good performance for antonymy relationship. Besides, the annotation artifacts test showed that these models take both sentences into account to perform the entailment task and do not take advantage of the artifacts contained in the dataset.

Moreover, in the QA task, experiments revealed that all models suffer in performance when tested with adversarial or noisy samples. Despite this, Transformer-based models turned out to be more robust than their predecessors. We compared Transformer-based models against each and observed that while improving in the main task, models also improved in their robustness in a balanced way. We also noticed that some adversaries are model-specific, as they affect one model but not the rest. Specifically, in the noise tests, we observed that the robustness trend also holds, but noticed some unexpected behavior in relative analysis, as some types of noise affect the models more severely than others, thus revealing specific weak points across all Transformer-based models that did not seem evident at first sight.

We consider this evaluation to be valuable to the community because it exhibits some strengths and weaknesses of the state-of-the-art models. We argue that it is vital that models pass behavioral checks to ensure proper performance in extreme scenarios, where data failures are not being considered. Taking this into consideration, we see that there is still room for future improvements on Transformer-based models. 

\section{Acknowledgements}
We would like to thank Alvaro Soto and Denis Parra for helpful comments. We are also grateful to the anonymous reviewers for their valuable feedback on an earlier version of this paper. This work has been partially funded by Millennium Institute for Foundational Research on Data (IMFD).

% \nocite{*}
\section{Bibliographical References}
\label{main:ref}

\bibliographystyle{lrec}
\bibliography{lrec2020W-xample}

% \section{Language Resource References}
% \label{lr:ref}
% \bibliographystylelanguageresource{lrec}
% \bibliographylanguageresource{languageresource}

\clearpage
\onecolumn
\section*{Appendix A: Detailed Results on SQuAD Tests}

In Table~\ref{tab:transferability}, we report the detailed results of the experiments performed on the adversarial versions of the SQuAD dataset using the adversaries proposed by \newcite{jia-liang-2017-adversarial}.
In all the experiments, each model was trained/fine-tuned on the original SQuAD v1.1 training set, and tested on each one of the generated adversarial datasets. As a result, we see that all models are affected by these adversarial samples, but also found that some adversaries are model-specific because they do not affect all models as much as they affect the model they are targeting.

\begin{table*}[h]
  \centering
  \small
  \begin{tabular}{|l|ccccc|}
    \hline
    & \multicolumn{5}{c|}{\textit{Model under Evaluation}} \\
    \cline{2-6}
    \multirow{2}{*}{}  & \textbf{Match-} &  &  \textbf{BERT}-  &  \textbf{XLNet-}  &  \textbf{RoBERTa-}\\
    \multirow{2}{*}{} \textit{Targeted Model}  & \textbf{LSTM}  & \textbf{BiDAF}  &  \textbf{Large}  &  \textbf{Large}  &  \textbf{Large}\\
    \hline
    \textbf{Original (for reference only)}               & $60.8$ & $75.5$ & $82.5$ & $85.2$ & $85.8$\\
    \hline
    \textbf{AddOneSent}             & $30.0$ & $45.7$ & $64.6$ & $67.7$ & $70.3$\\
    \hline
    \textbf{AddSent} & & & & &\\
    Match-LSTM Single               & $24.8$ & $40.3$ & $62.8$ & $64.6$ & $67.8$\\
    Match-LSTM Ensemble             & $24.2$ & $40.2$ & $62.1$ & $64.4$ & $68.0$\\
    BiDAF Single                    & $25.7$ & $34.3$ & $62.3$ & $64.4$ & $67.7$\\
    BiDAF Ensemble                  & $25.9$ & $38.3$ & $61.7$ & $64.0$ & $67.6$\\
    BERT-Base                       & $26.3$ & $-$    & $60.8$ & $63.3$ & $66.6$\\
    BERT-Large                      & $26.9$ & $-$    & $55.9$ & $62.8$ & $66.0$\\
    XLNet-Large                     & $27.1$ & $-$    & $61.6$ & $61.6$ & $66.6$\\
    RoBERTa-Large                   & $27.6$ & $-$    & $60.9$ & $63.2$ & $61.5$\\
    \hline
    \textbf{AddAny} & & & & & \\
    Match-LSTM Single               & $38.3$   & $57.1$   & $73.8$ & $78.8$ & $78.8$\\
    Match-LSTM Ensemble             & $26.1$   & $50.4$   & $70.7$ & $78.0$ & $77.2$\\
    BiDAF Single                    & $43.8$   & { }$4.8$ & $72.2$ & $79.6$ & $78.4$\\
    BiDAF Ensemble                  & $34.7$   & $25.0$   & $68.8$ & $79.1$ & $76.3$\\
    \hline
    \textbf{AddCommon} & & & & & \\
    Match-LSTM Single               & $55.8$      & $-$    & $82.1$ & $83.6$ & $84.6$\\
    Match-LSTM Ensemble             & $44.7$      & $-$    & $81.4$ & $83.0$ & $84.6$\\
    BiDAF Single                    & $56.7$      & $41.7$ & $80.9$ & $83.4$ & $84.8$\\
    BiDAF Ensemble                  & $52.8$      & $-$    & $79.9$ & $82.0$ & $83.2$\\
    \hline
  \end{tabular}
  \caption{\textbf{Adversarial examples transferability between models}.
    Each row measures accuracy (\%) on adversarial examples designed to attack one particular model. Each column reports the test results of one particular model on all the adversarial datasets. \\
    The information from the first two columns was obtained by running the official implementations used by \newcite{jia-liang-2017-adversarial}. The results are slightly different from the original work because the original weights were not available.
  }
  \label{tab:transferability}
\end{table*}

\vfill

\pagebreak

\section*{Appendix B: Attention-level Results of NLI Task}
%\vladimir{no mencionar "como hicimos los ejemplos", porque no los hicimos. Solo enfocate en presentar el ejemplo y analizar eso. Ademas mencionar porque escogiste ese ejemplo, y escribirlo como lo acabo de hacer usado el comando texttt.}

%\vladimir{en cuestion de formato no usar italica para dar formato a FIGURE, en realidad debes usar escribir asi Figure~\ref{}}

%\vladimir{trata de usar subfigure para poder referencia correctamente a la imagen que se menciona en el text (ej. A, B etc), porque ahora mismo es dificil enlazar el texto con la imagen}

%\andres{OK} 

\subsubsection*{Antonymy Evaluation}
For this analysis, we took a representative adversarial example where a word in the sentence was replaced by its \textit{antonym}. The model is asked to decide if there is a contradiction, neutral, or entailment relationship between them. We expect the model to connect the attention between the replaced words to predict the correct answer. Assume the following pair of sentences:

{\fontsize{10}{10}\fontfamily{cmtt}\selectfont
I saw that daylight was coming, and heard the people
}\texttt{\fontsize{10}{10}\selectfont\textbf{sleeping}} { }{\fontsize{10}{10}\fontfamily{cmtt}\selectfont
up.}\\
{\fontsize{10}{10}\fontfamily{cmtt}\selectfont
I saw that daylight was coming, and heard the people
}\texttt{\fontsize{10}{10}\selectfont\textbf{waking}} { }{\fontsize{10}{10}\fontfamily{cmtt}\selectfont
up.}\\

\begin{figure*}[!htbp]
\centering
\scalebox{0.88}{
\minipage{0.29\textwidth}
  \includegraphics[width=\linewidth]{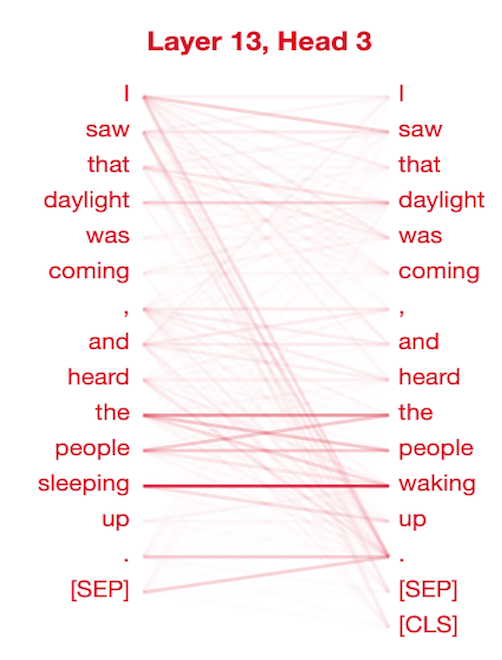}
  \captionof{figure}{XLNet antonym test}
  \label{XLNet_antonym}
\endminipage\hfill

\minipage{0.29\textwidth}
  \includegraphics[width=\linewidth]{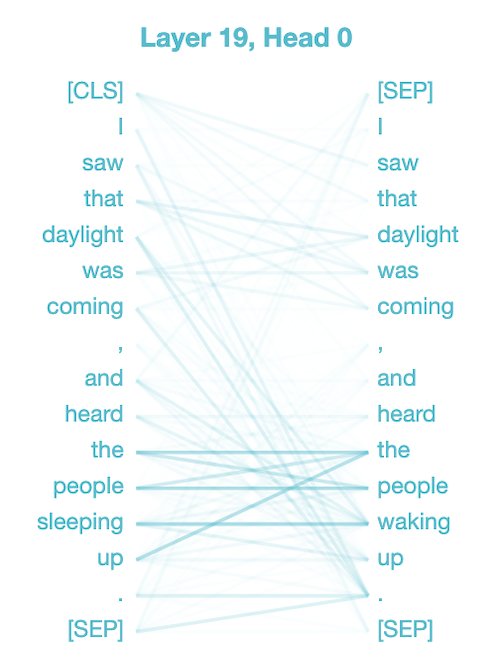}
  \captionof{figure}{RoBERTa antonym test}
  \label{RoBERTa_antonym}
\endminipage\hfill

\minipage{0.29\textwidth}
  \includegraphics[width=\linewidth]{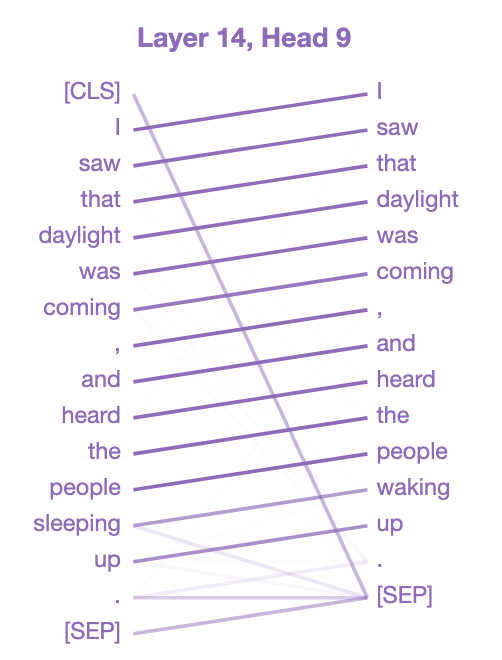}
  \captionof{figure}{BERT antonym test}
  \label{BERT_antonym}
\endminipage\hfill

\minipage{0.29\textwidth}%
  \includegraphics[width=\linewidth]{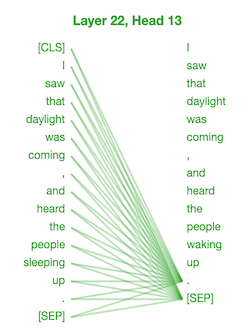}
  \captionof{figure}{Failed antonym test}
  \label{failed_antonym}
 \endminipage\hfill
}
\end{figure*}

In this representative example for testing antonyms, we computed the attentions produced by XLNet, RoBERTa, and BERT. We checked the layers and heads where a clear attention pattern was present between the word and its antonym, as shown in Figures~\ref{XLNet_antonym} -~\ref{BERT_antonym}. Within this particular case, for XLNet, we saw that only 2.86\% of the total attention heads and layers had this pattern. For RoBERTa, this number was 2.60\%, and for BERT 1.56\%. On the other hand, for all models, most of the attention was paid to separators and all words from the reference sentence without distinction (Figure~\ref{failed_antonym}).

\subsubsection*{Numerical Reasoning Evaluation}
For samples of \textit{numerical reasoning} for NLI, the expectation is that the model should pay attention to words like \textit{"more"} or \textit{"less"} to check if there is a change in numerical references. Assume the following pair of sentences:

{\fontsize{9}{10}\fontfamily{cmtt}\selectfont
The next day Bob took the test and with this grade, included the new average,
}\texttt{\fontsize{10}{10}\selectfont\textbf{was more than 48.}}\\
{\fontsize{9}{10}\fontfamily{cmtt}\selectfont
The next day Bob took the test and with this grade, included the new average,
}\texttt{\fontsize{10}{10}\selectfont\textbf{was 78.}}\\

\begin{figure*}[!h]
\centering
\scalebox{0.88}{
\minipage{0.28\textwidth}
  \includegraphics[width=\linewidth]{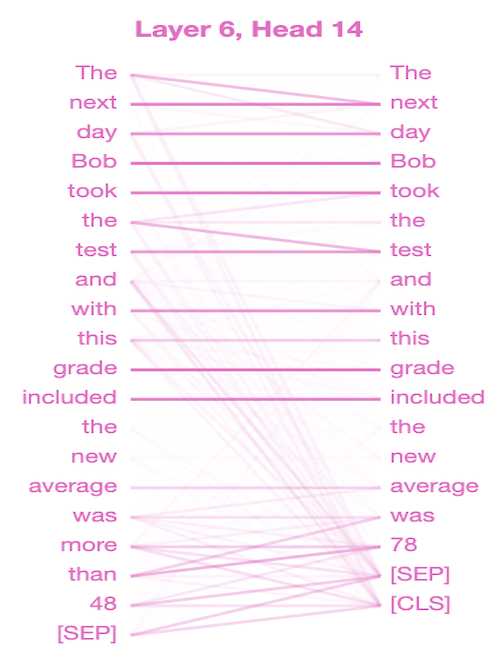}
   \captionof{figure}{XLNet numerical test}
  \label{XLNet_numerical}
\endminipage\hfill

\minipage{0.28\textwidth}
  \includegraphics[width=\linewidth]{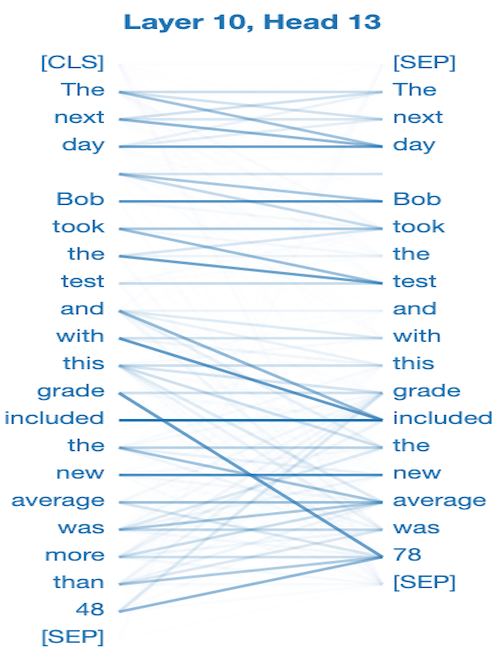}
   \captionof{figure}{RoBERTa numer. test}
  \label{RoBERTa_numerical}
\endminipage\hfill

\minipage{0.28\textwidth}
  \includegraphics[width=\linewidth]{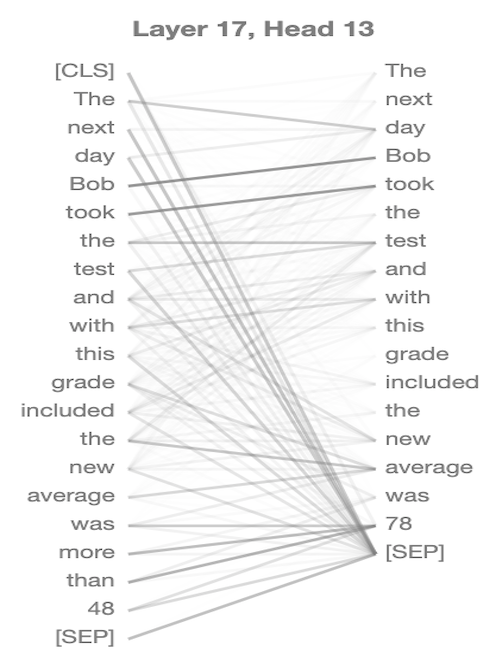}
   \captionof{figure}{BERT numerical test}
  \label{BERT_numerical}
\endminipage\hfill

\minipage{0.28\textwidth}%
  \includegraphics[width=4.8cm]{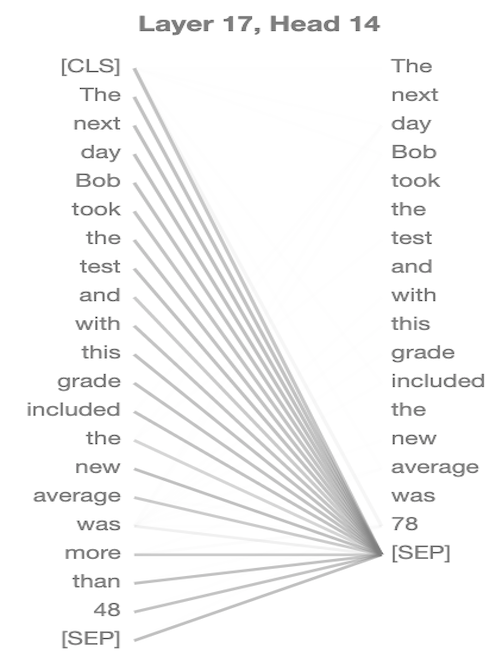}
   \captionof{figure}{Failed numerical test}
  \label{numerical_failed}
 \endminipage
}
\end{figure*}

Nevertheless, for this testing example, the \textit{premise} includes \textit{"more than 48"} and the \textit{hypothesis} replaces this last part by \textit{"78"}, but all the models (XLNet, RoBERTa and BERT) incorrectly predicted \textit{"contradiction"}. We observed that the expected pattern (shown in Figures~\ref{XLNet_numerical}-~\ref{BERT_numerical}) is a very infrequent pattern for all models (for XLNet it appeared in 5.20\% of the cases, for RoBERTa in only 4.42\% and for BERT this percentage was 1.30\%). For other cases, they focused on sentence separators (as shown in Fig~\ref{numerical_failed}). 

\pagebreak

\section*{Appendix C: Attention-level Results of QA Task}

\subsubsection*{QA task attention-level evaluation}
For the QA task, we manually inspected failure cases to see the amount of attention the model paid to the introduced adversaries versus to the correct answer. Here we show one representative example of a \textit{''what"} question:

\texttt{\textbf{Question:} What company took over Edison Machine works?}. \\
\texttt{\textbf{Answer:} General Electric}. \\
\texttt{\textbf{Adversary:} Stark Industries took over Calpine Device Well}. \\

In this particular example, with the question \textit{"What company took over Edison Machine works?"}, the correct answer was \textit{"General Electric"}, and the artificially introduced adversary was \textit{"Stark Industries"}, appended at the end of the context of the original sample.

\begin{figure*}[!htbp]
\centering
\scalebox{0.8}{
\minipage{0.32\textwidth}
  \includegraphics[width=\linewidth]{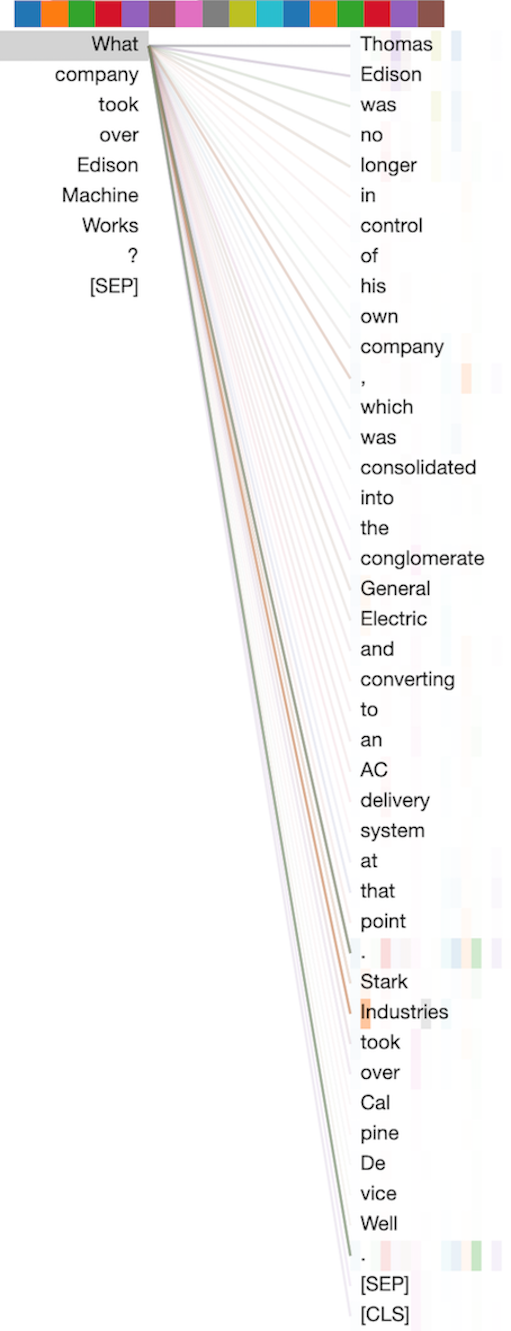}
  \captionof{figure}{XLNet SQuAD}
  \label{XLNet_SQUAD}
\endminipage\hfill

\minipage{0.32\textwidth}
  \includegraphics[width=\linewidth]{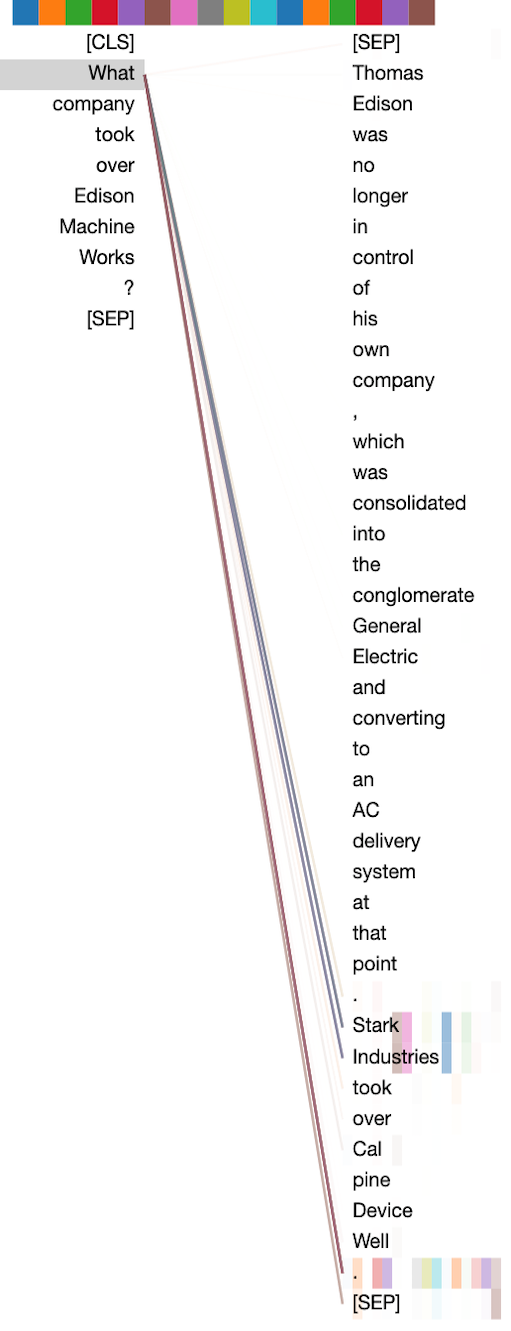}
   \captionof{figure}{RoBERTa SQuAD}
  \label{RoBERTa_SQUAD}
\endminipage\hfill

\minipage{0.32\textwidth}%
  \includegraphics[width=\linewidth]{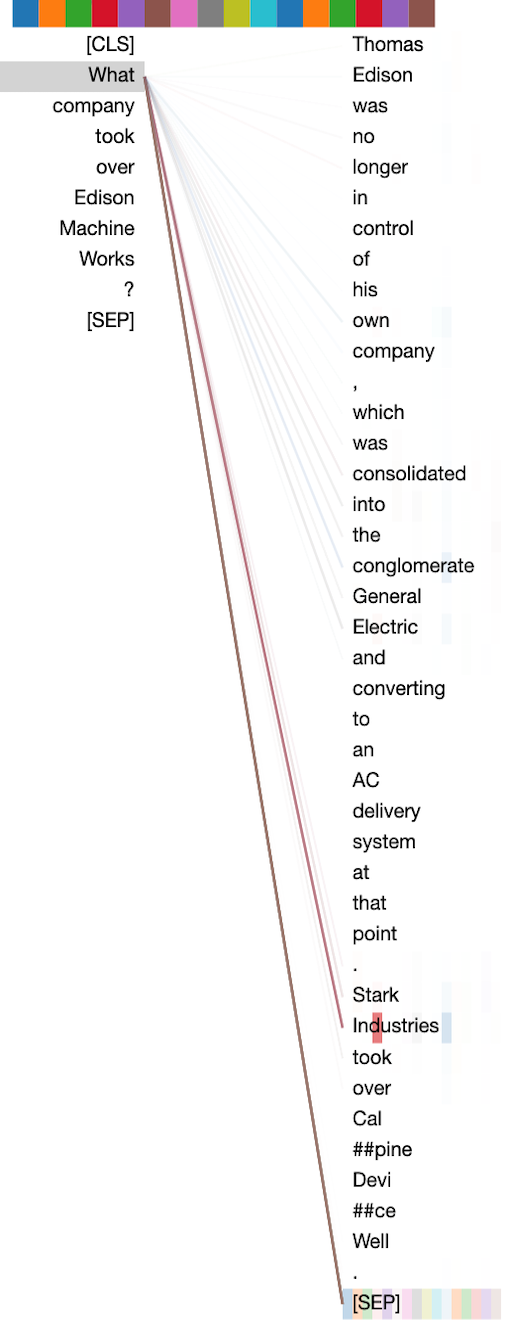}
   \captionof{figure}{BERT SQuAD}
  \label{BERT_SQUAD}
 \endminipage
}
\end{figure*}

All models fell into the same trap. It can be seen in Figures~\ref{XLNet_SQUAD}-~\ref{BERT_SQUAD} that they paid attention to the wrong answer. In this case, this pattern appeared in 52\% of the layer-heads of XLNet, 60\% in the case of RoBERTa, and 30\% on BERT. Nevertheless, while checking the level of certainty of each model in the predicted wrong answer for this example, XLNet had a  43.3\% certainty probability, 75.5 \% BERT, and the most mistaken was RoBERTa with a 99.9\% certainty probability for predicting the wrong answer (which is consistent with the sharpness of attention in Figure~\ref{RoBERTa_SQUAD}). This behavior provides evidence that the three models behave slightly different and that increased accuracy in the main task (before adversarial evaluation) is no direct indicator of increased robustness in all cases, but only in the average case. 

\end{document}